\documentclass{ifacconf}

\usepackage{graphicx}      % include this line if your document contains figures
\usepackage{natbib}        % required for bibliography
\usepackage{dirtytalk}
\usepackage{subfigure}
\usepackage{wrapfig}
\usepackage{amsmath}
\usepackage{footnote}
\begin{document}
\begin{frontmatter}

\title{A Supervised Machine Learning Approach to Operator Intent Recognition for Teleoperated Mobile Robot Navigation\thanksref{footnoteinfo}} 

\thanks[footnoteinfo]{was supported by the UKRI-EPSRC grant EP/R02572X/1 (UK National Centre for Nuclear Robotics).}

\author[First]{Evangelos Tsagkournis} 
\author[First]{Dimitris Panagopoulos}
\author[Second]{Giannis Petousakis}
\author[First]{Grigoris Nikolaou}
\author[Third]{Rustam Stolkin}
\author[Third]{Manolis Chiou}

\address[First]{Industrial Design \& Production Engineering,            University of West Attica, Athens, Greece (e-mail: \{auto45543, auto44990, nikolaou\}@uniwa.gr)}
\address[Second]{Cognitive Robotics Lab, 
   University of Manchester, Manchester, UK (e-mail: ioannis.petousakis@postgrad.manchester.ac.uk)}
\address[Third]{Extreme Robotics Lab (ERL), 
   University of Birmingham, Birmingham, UK, (e-mail: \{r.stolkin, m.chiou\}@bham.ac.uk)}

\begin{abstract}                % Abstract of not more than 250 words.
In applications that involve human-robot interaction (HRI), human-robot teaming (HRT), and cooperative human-machine systems, the inference of the human partner's intent is of critical importance. This paper presents a method for the inference of the human operator's navigational intent, in the context of mobile robots that provide full or partial (e.g., shared control) teleoperation. We propose the Machine Learning Operator Intent Inference (MLOII) method, which a) processes spatial data collected by the robot's sensors; b) utilizes a supervised machine learning algorithm to estimate the operator's most probable navigational goal online. The proposed method's ability to reliably and efficiently infer the intent of the human operator is experimentally evaluated in realistically simulated exploration and remote inspection scenarios. The results in terms of accuracy and uncertainty indicate that the proposed method is comparable to another state-of-the-art method found in the literature.
\end{abstract}

\begin{keyword}
Human-Robot Interaction, Human-Robot Teaming, Human-in-the-loop, Operator Navigational Intent, Supervised Machine Learning, Random Forest.
\end{keyword}

\end{frontmatter}
%===============================================================================

\section{Introduction}
% general intro
Robotic systems, which are capable of operating alongside humans in public environments are becoming increasingly relevant and essential to modern applications. Such human-robot systems (HRS), are starting to be widely deployed in real world scenarios including search and rescue, area inspection, defence, and disaster response.

% general about why we do need modules like that
When it comes to HRT, discerning the human state (e.g., human intent \cite{Jain2019} or attention \cite{Petousakis2020}) has a beneficial effect on making the interaction between human and robot more efficient and reliable \cite{Murphy}. In order to facilitate collaboration towards successful HR team performance \cite{Liu2016}, the AI agent has to be able to infer its human partner's intent. Correctly inferring the human intent holds immense potential for applications such as assistive robotics \cite{Demiris2007} and variable autonomy systems \cite{pappas, chiouMI, Nemics}, as it allows the AI agent to adapt its policies based on the information received. 

This is especially true for Mixed-Initiative (MI) systems, in which the two agents (i.e., human operator and AI agent) hold equal authority on switching between the various Levels of Autonomy (LOA) during task execution \cite{Jiang2015}. The performance of the Human-Robot Team (HRT) can be improved by introducing human intent inference in an effort to mitigate adverse effects such as conflict for control \cite{hier}. Previous literature \cite{chiouMI} has identified conflict for control typically occurring when the operator cannot explicitly communicate their intent. As a result, the AI agent's attempt to switch the LOA sets off a cycle in which both agents repeatedly override each other's commands, while operating a mobile robot in remote disaster response scenarios.

% specific effects made by intent on conflict 
%Some specific examples of applications which can benefit from Correctly inferring the operator's intent holds immense potential for assistive robotics \cite{Demiris2007} and can be leveraged by variable autonomy systems such as shared control \cite{pappas} or Mixed-Initiative (MI) \cite{chiouMI, Nemics} systems. This is especially true for MI systems, whose performance can be significantly improved, when incorporating human intent recognition along with other intelligent functionalities in an effort to mitigate conflict for control \cite{hier}. 

In this paper, we propose the Machine Learning Operator Intent Inference (MLOII), a method for the classification of human intent. We use it to predict the human operator's current navigational goal (i.e., waypoint) when a robot is fully or partially teleoperated. Our approach leverages an offline learning phase in which a Random Forest classifier is trained based on the mobile robot's spatial properties; (i) the approach velocity; (ii) the Euclidean distance; (iii) and the orientation angle relative to the potential navigational goals. During run-time, the trained MLOII is deployed in order to predict the goal the human operator tries to approach. The contribution of this paper is threefold. First, we employ a data-driven, machine learning, approach to infer the operator's intent in the context of robot teleoperation. Second, we implement and analyse the MLOII in multiple high-fidelity simulated remote inspection scenarios. Third, we provide evidence for the feasibility (i.e., proof-of-concept) of our proposed approach by comparing the MLOII to another method \cite{BOIR} found in the literature.

% We contribute a ML approach based on X to teleopeated mobile robot navigarion..
% We compare our approach to a recent intent inference approach basd on bayes.... which was shown to outperform X and Y methods
% "We provide evidence for the feasibility of our proposed approach"

\section{Related Work}
Accurately predicting the human intent from the robot's standpoint \cite{Demiris2007} is important for safe and efficient robot navigation and interaction with humans. In support of this endeavor, a wide variety of literature exists.

% confidence functions
A confidence function has been proposed in \cite{Carlson2008} that blends the output from two exponential decay models in order to estimate the user's intent in assistive wheelchair navigation tasks. Similarly, the method in \cite{Dragan2013} merges instantaneous observations, such as proximity to the target and user or robot generated commands, to evaluate a confidence function while teleoperating a robotic manipulator.

% bayesian statistics
Other approaches have focused on developing Bayesian probabilistic methods to estimate human intent across different tasks. For instance, the work in \cite{Huntemann2013} allows the robot agent to capture robotic wheelchair users' intent during shared control navigation. Similarly, in \cite{Jain2019} the intent could be derived as a means to assist manipulation tasks (i.e., the object humans want to engage with). Both works fuse present and past local information with a user-specific model to reason about the underlying human intent. When performing teleoperated navigation in safety-critical applications, the work in \cite{BOIR} on intent recognition, has been found to contribute significantly in reducing conflicts between human and AI in MI systems \cite{hier}.

% learning techniques
In addition to these techniques, many works leverage learning methods. In particular, a predictive neural network is proposed in \cite{YanLSTM} to determine human intended movement using skeleton-based motion information. Several machine learning algorithms are compared in \cite{CHOI}, in an effort to predict the intent of drivers from surrounding vehicles to lane-changing scenarios. In \cite{PETKOVIC2019182} a hidden Markov model (HMM) framework is used to estimate the goal of the human in an automated warehouse. Finally, a robotic wheelchair that uses disentangled Variational sequence encoder for inferring a discrete intent variable from human behavior by clustering action plans is presented in \cite{zolotas2021}.

To the best of our knowledge most operator's intent published research is found in the fields of assistive robotics and autonomous driving. This paper aims to focus its contribution on the field of remotely operated mobile robots. Our proposed learning method allows classification of navigational intent by making use of real time sensor data. It also presents an opportunity for further development by adding contextual information (i.e., hazards, victims) to the system, enabling for more challenging scenarios during evaluation. Finally, our method can also be used in scenarios in which a human operator is physically present with the robot.

%The choice of this type of learning lies in the nature of the data that since a set of tagged training is used, that is, the origin of the data is known.

\section{Machine Learning Operator Intent Inference - MLOII} \label{new_approach}
% REVISED version of the above based on comments
\subsection{Problem Formulation}
\label{problem_formulation}
In this section, we introduce the MLOII method, which enables an AI agent to infer the human operator's intent in HRI schemes based on machine learning Random Forest algorithm. Drawing inspiration from the field of safety-critical applications, the problem is formulated as a classification task of the potential navigational goals (i.e., waypoints) in the context of a remote area exploration. In particular, the human-robot team is tasked with exploring an area and closely inspecting certain points-of-interest (POIs) for further examination \cite{Rognon2018}. We define a discrete set $\textbf{\textit{G}} = \big\{g^{1}, g^{2} ,\dots, g^{N} \big\}$ containing the navigational goals, where $N$ is the total number of goals corresponding to each POI in the area. Here, we assume that the navigational goals, whose locations have been specified in advance, are represented by the POI. In practice, these goals can be predefined by a UAV that monitors and updates the current map status, ensuring sufficient area coverage \cite{Chatziparaschis}. Although this paper focuses on exploration performed by ground robots, our ideas may be applicable to other tasks and types of robots.

\subsection{Data Preprocessing} \label{data_pre}
To generate a dataset for training and testing the MLOII method, we developed two different simulated environments with fixed locations for each goal. The starting position and orientation of the robot were randomized in each trial to introduce variability. During offline testing, a human operator navigated the robot from the starting point to a randomly selected goal, while the onboard sensors recorded data labeled according to the trial's specific goal. After labeling, the data was further processed to extract the relevant features for training the model. The three types of derived features used to train the MLOII are:

\begin{itemize}
    \item The approach velocity $\nu$ (i.e., the vector quantity that relates to the rate at which the mobile robot alters its position as it advances towards each goal).
    \item The Euclidean distance $d$ between the mobile robot and each goal.
    \item The angle \(\theta\) (i.e., the mobile robot's orientation with respect to each goal in the environment).
\end{itemize}

The dataset was carefully balanced to ensure that each label had an equal number of instances, thereby preventing any bias in the learning algorithm. Finally, the dataset was split into two subsets: a training set and a testing set, with $70\%$ and $30\%$ of the original dataset allocated to each one respectively.

\subsection{Supervised Learning: Random Forest} \label{RFC}
We used the Random forest algorithm \cite{Breiman2001}, which is a supervised machine learning algorithm widely used in both regression and classification problems. It is an ensemble method, growing a number of decision trees, with each tree being trained on a subset sampled from the training data uniformly and with replacement. Therefore, some observations may be repeated in each subset. Ensemble simply means that the method is made up of a set of classifiers or regressors (e.g., decision trees) and their predictions are aggregated to identify the most popular result. In \cite{https://doi.org/10.48550/arxiv.2207.08815}, tree-based models are found to perform well on medium-sized tabular data. The classification procedure involves the estimation of the input-output mapping function $\mathbf{y = f(x)}$. $\mathbf{x}$ is the input feature set from empirical data and $\mathbf{y}$ is the corresponding output variable, which in our case is the human operator's predicted goal . More specifically the output of the classifier is the class label it predicts. The confidence for the prediction of each class label is the ratio of the number of decision trees that predicted the particular class to the total number of decision trees available to the classifier. Since the MLOII is intended to use the onboard processing power of the robot with almost real time constraints while demanding high accuracy in its predictions, it was configured to use 50 decision trees. This configuration allowed for reasonable classification accuracy (CA) while being computationally inexpensive.

\section{Experimental Evaluation}
In this section, we provide details on the experimental validation of the proposed method described in Section \ref{new_approach}. We quantitatively evaluate its effectiveness over a set of multiple experiments, inspired by disaster response and remote inspection scenarios.

% Commented out and replaced with the one below
\subsection{Apparatus \& Software}
% In the experiments conducted, we used a realistically simulated mobile robot. The mobile robotic platform chosen was Husky, a medium-sized unmanned ground vehicle from Clearpath Robotics. The simulated mobile robot was equipped with a laser range finder designed for autonomous mobile systems and used for localization, navigation, and detection purposes. Along with the simulated robotic platform, our experimental setup also consists of an Operator Control Unit (OCU) that allows communication between a human operator and the robot. The OCU consists of a mouse and a joystick as input devices, a monitor screen displaying the Graphical User Interface (GUI) and a laptop running the simulated environment (see Fig. \ref{fig:OPERATOR_UNIT}). The simulation took place in a high-fidelity 3D robotics simulator called Gazebo, offering a rich environment to quickly develop, deploy, and test our method. The software and related functionalities were developed in the Robot Operating System (ROS).

%WIP paraphrasing for the Apparatus & Software subsection

\begin{figure}
	\centerline{\subfigure[]{\includegraphics[width=0.48\columnwidth]{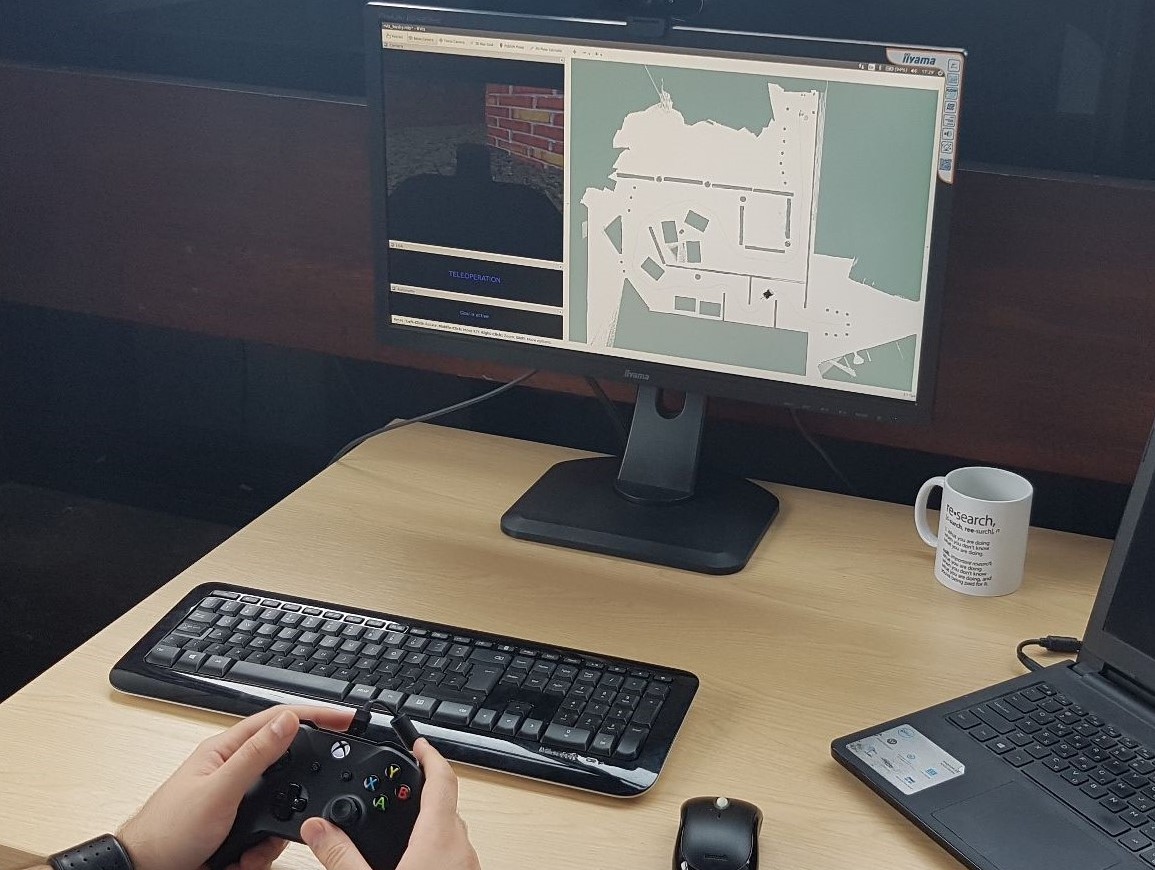}
			\label{fig:OCU}}
		\hfil
		\subfigure[]{\includegraphics[width=0.4\columnwidth]{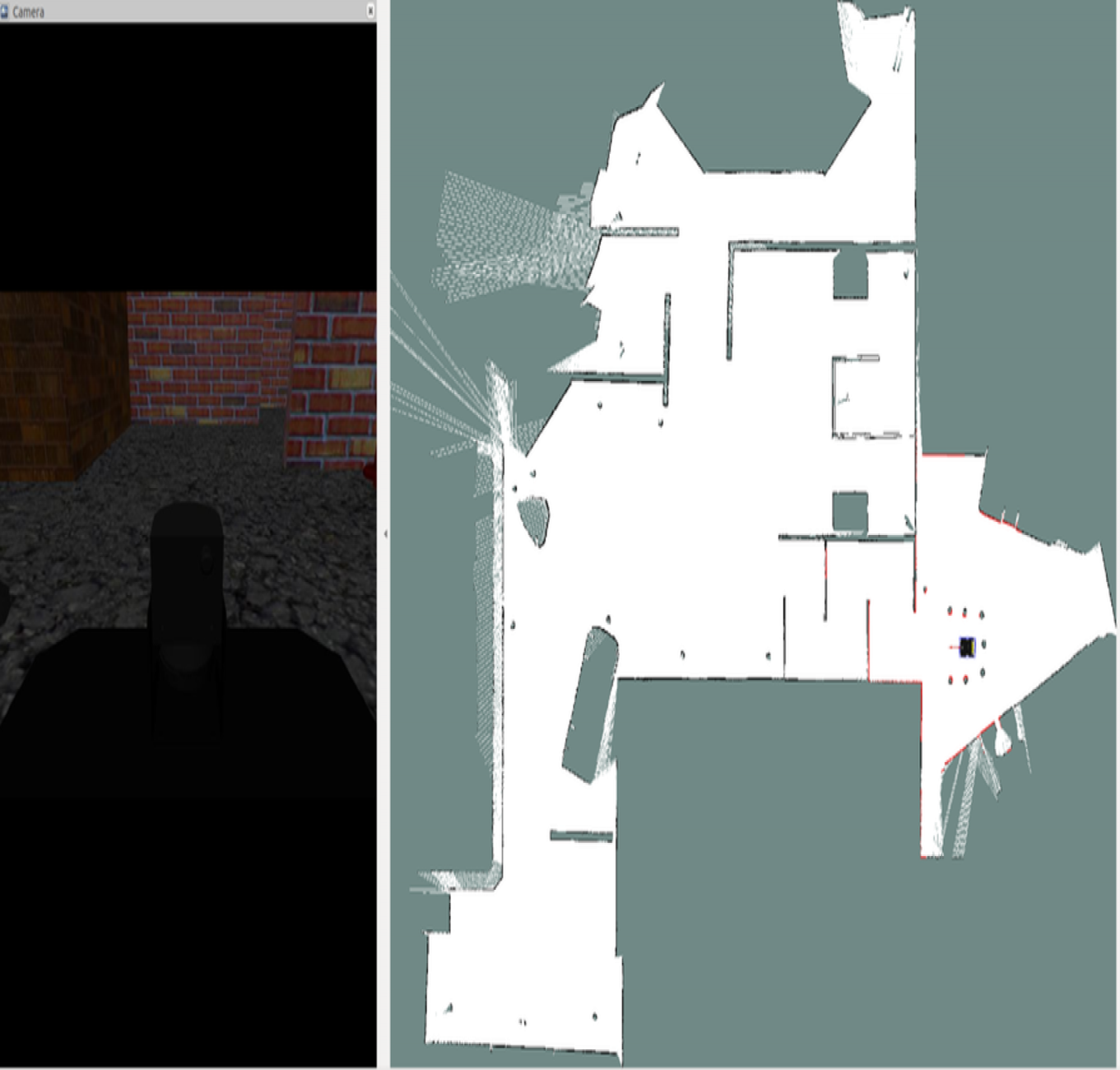}
			\label{fig:GUI}}}
	\caption{\textbf{\ref{fig:OCU}:} The Operator Control Unit (OCU): composed of a mouse and a joystick, a laptop and a screen showing the Graphical User Interface (GUI) and the simulation environment. {\textbf{\ref{fig:GUI}}:} The Graphical User Interface (GUI) and the simulation environment with the video feed from the camera.}
	\label{fig:OPERATOR_UNIT}
\end{figure}

The experiments were carried out using a simulated Husky mobile robot, a medium-sized robotic platform by Clearpath Robotics. The robot was equipped with a laser range finder that was designed for autonomous mobile systems, which was used for localization, navigation, and detection purposes. In addition to the simulated robot platform, an Operator Control Unit (OCU) was used to facilitate communication between the human operator and the robot. The OCU comprised a mouse and a joystick as input devices, a monitor screen displaying the Graphical User Interface (GUI), and a laptop that ran the simulated environment (see Fig. \ref{fig:OPERATOR_UNIT}). The simulation was performed using Gazebo, a high-fidelity 3D robotics simulator that provided a rich environment for the development, deployment, and testing of the proposed method. The software and related functionalities were developed using the Robot Operating System (ROS) environment. The experimental setup was designed to mimic disaster response and remote inspection scenarios, and multiple experiments were conducted to quantitatively evaluate the effectiveness of the proposed method.

% \begin{figure}
% 	\centering
% 	\includegraphics[width=0.55\columnwidth]{img/ocu1.jpg}
% 	\caption{The Operator Control Unit (OCU): composed of a mouse and a joystick, a laptop and a screen showing the Graphical User Interface (GUI) and the simulation environment}
% 	\label{fig:OCU}
% \end{figure}

% \begin{figure}
% 	\centering
% 	\includegraphics[width=0.99\columnwidth]{img/ocu1.jpg}
% 	\caption{The Operator Control Unit (OCU): composed of a mouse and a joystick, a laptop and a screen showing the Graphical User Interface (GUI) and the simulation environment.}
% 	\label{fig:OCU}
% \end{figure}

\begin{figure}
	\centerline{\subfigure[]{\includegraphics[width=0.48\columnwidth]{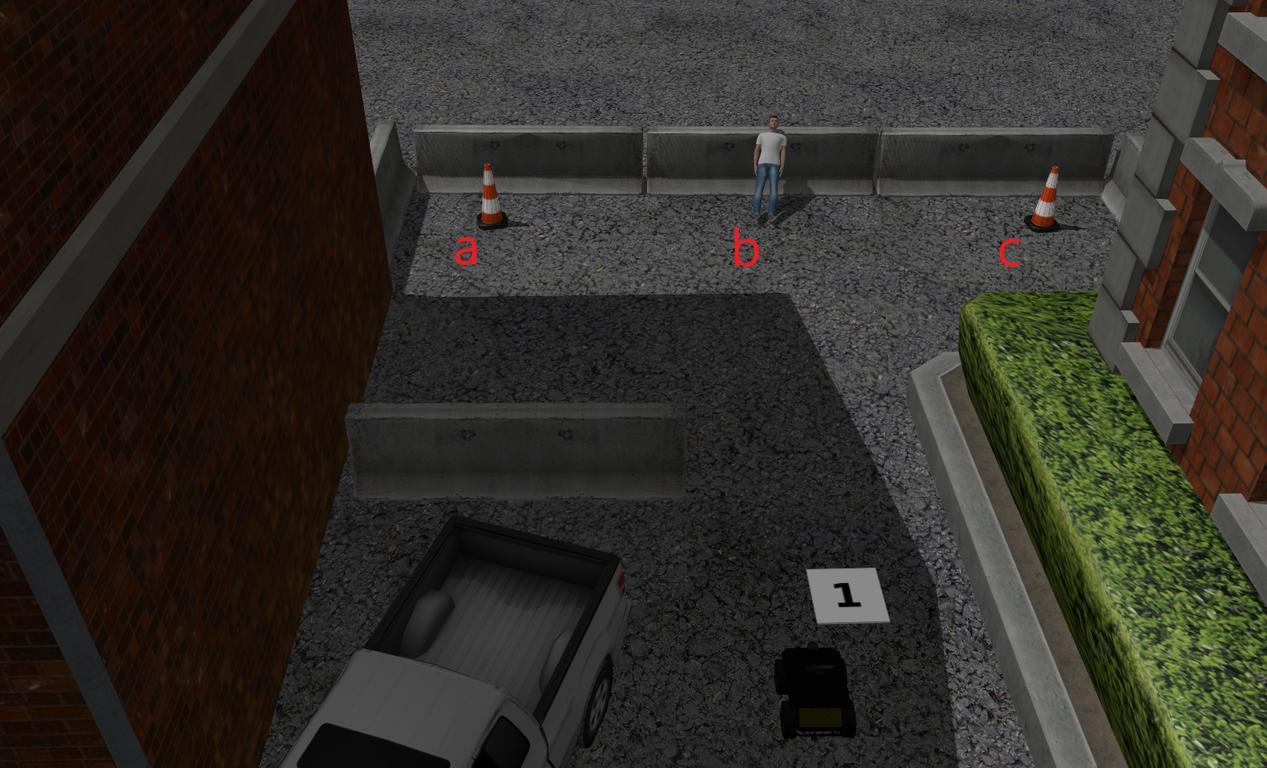}
			\label{fig:area1}}
		\hfil
		\subfigure[]{\includegraphics[width=0.48\columnwidth]{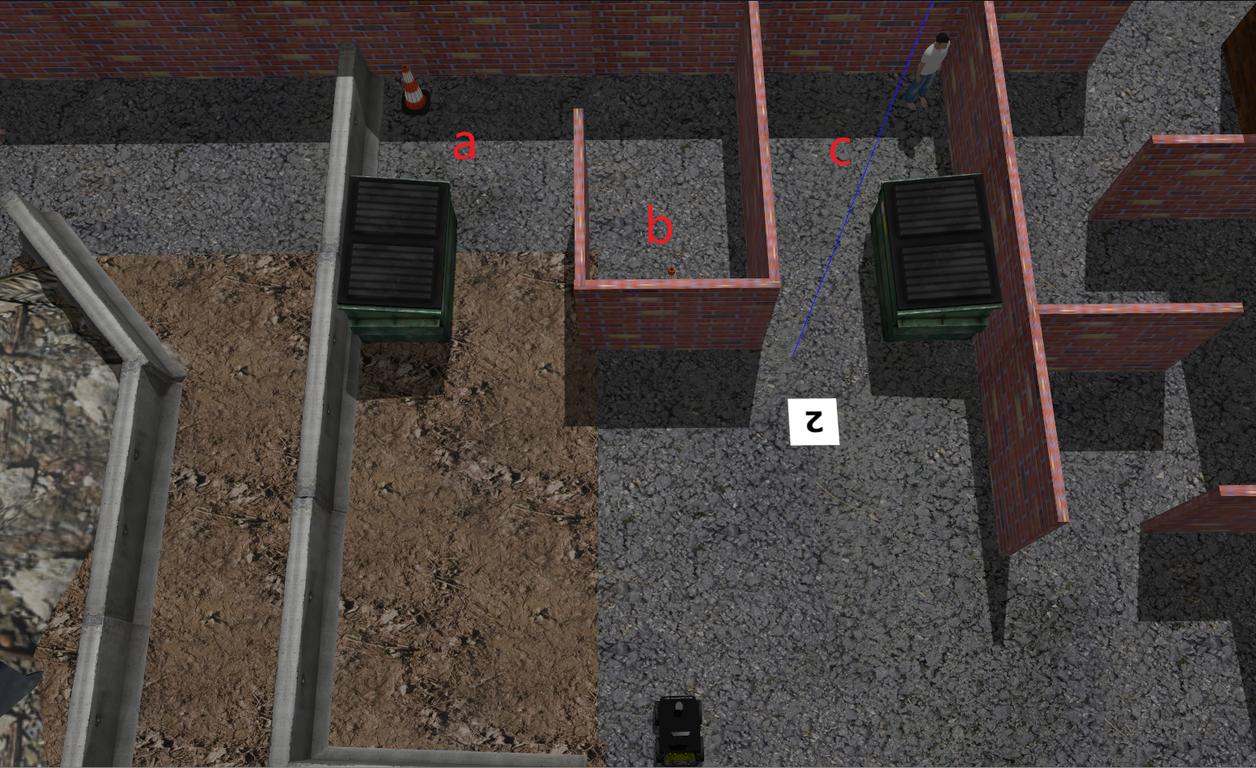}
			\label{fig:area2}}}
		\hfil
		\subfigure[]{\includegraphics[width=0.48\columnwidth]{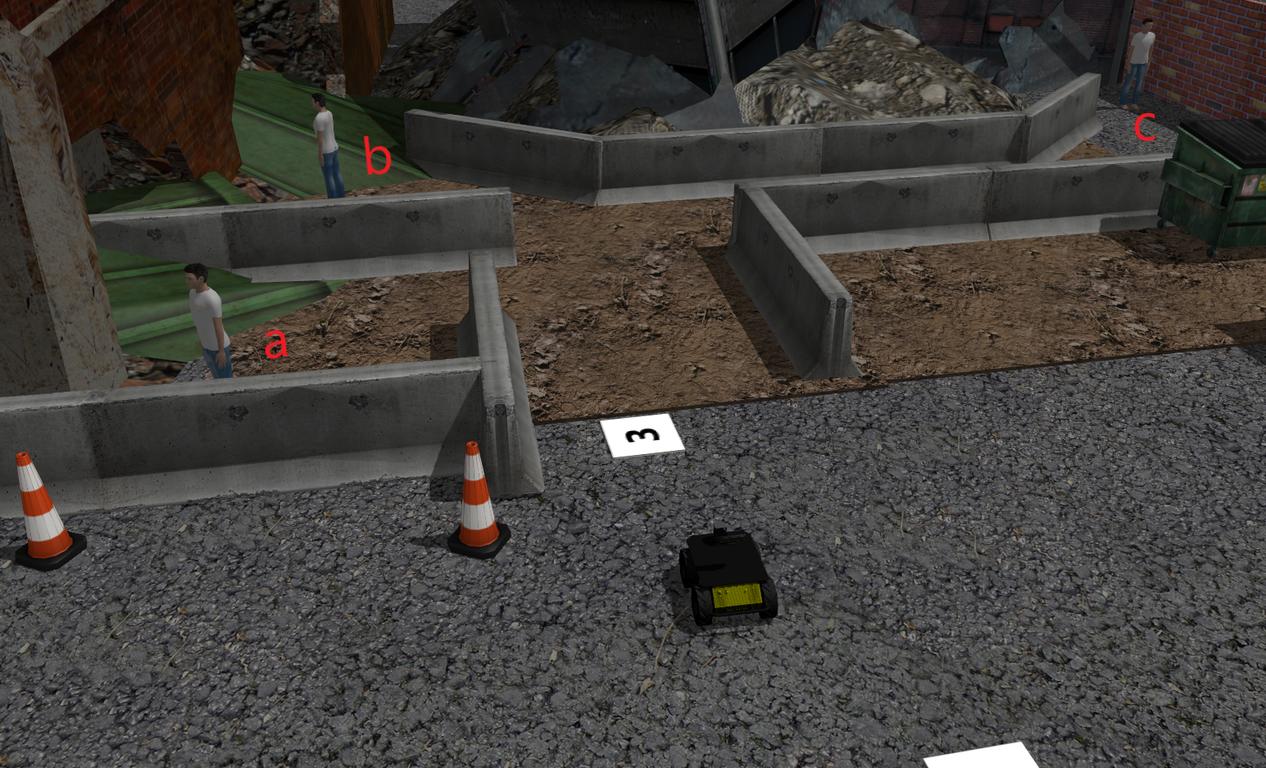}
			\label{fig:area3}}
		\subfigure[]{\includegraphics[width=0.48\columnwidth]{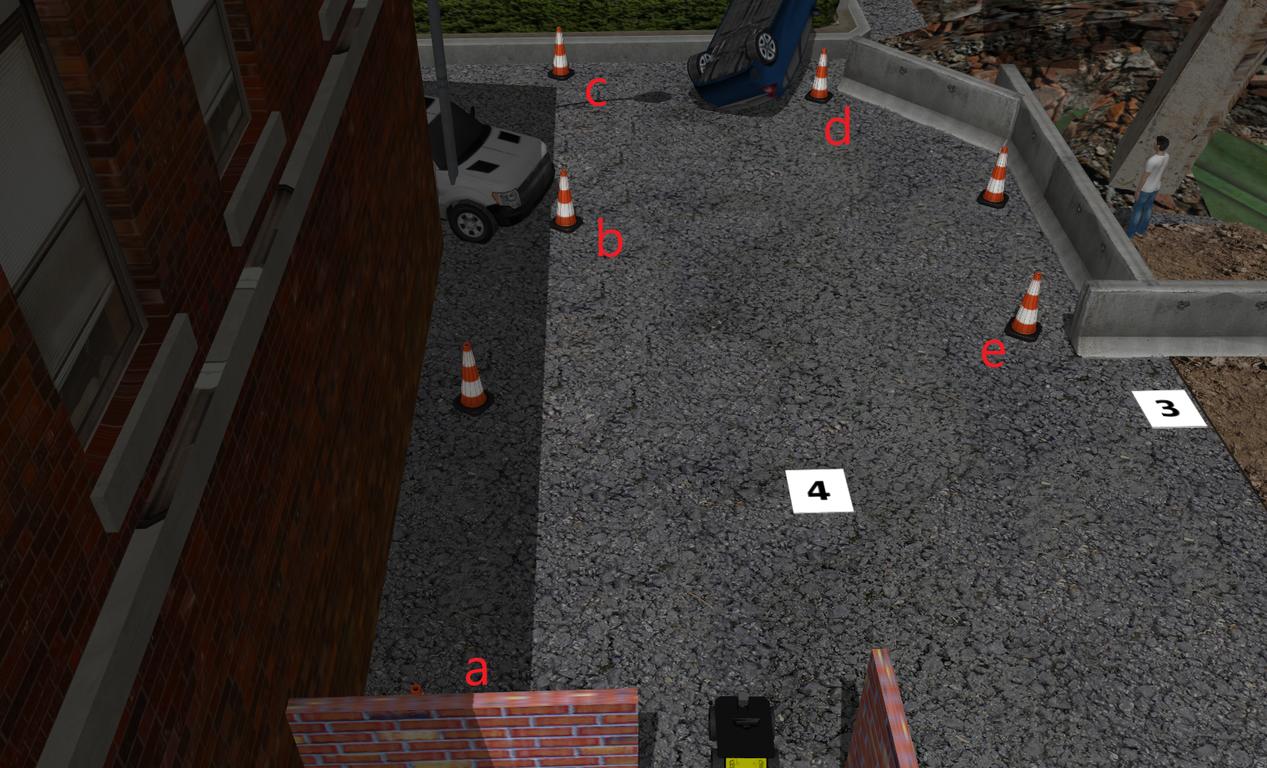}
			\label{fig:area4}}
% 		\subfigure[]{\includegraphics[width=0.48\columnwidth]{img/ocu1.jpg}
% 			\label{fig:OCU}}
	\caption{\textbf{\ref{fig:area1}}-\textbf{\ref{fig:area4}:} The evaluation scenarios. The numbers denote the number of specific scenario; the red lower-case letters denote the points-of-interest for exploration (i.e., the set of possible goals); the point in which the husky robot stands refers to the starting point of each scenario; the humans denote the intended goals.} %{\textbf{\ref{fig:OCU}}:} The Operator Control Unit (OCU): composed of a mouse and a joystick, a laptop and a screen showing the Graphical User Interface (GUI) and the simulation environment.}
	\label{fig:scenarios}
\end{figure}

\begin{table*}[t]
    \caption{Descriptive Statistics (Accuracy and Log-Loss) \& Pairwise comparisons.}
	\centering
	
	    \begin{tabular}{lllllll}
		\hline
		\textbf{Scenarios}&\textbf{descriptive}& & & \textbf{descriptive}&  & \\
	    
	    \textbf{\&}& \textbf{statistics}& \textbf{test} &\textbf{p-value} &\textbf{statistics} & \textbf{test} &\textbf{p-value} \\
	    
	    \textbf{Methods}&\textbf{(accuracy-\%)}&\textbf{statistic} &\textbf{(accuracy)}&\textbf{(log-loss)}& \textbf{statistic} &\textbf{(log-loss)} 
		
		\\ \hline
		\begin{tabular}[c]{@{}l@{}}  \textit{Scenario 1} \\ MLOII \\ BOIR \\ MLOII vs BOIR \end{tabular}
		&
		\begin{tabular}[c]{@{}l@{}}\\$M = 95.60, SD = 2.29$ \\ $M = 85.30, SD = 3.83$ \\ $-$ \end{tabular}
		& 
		\begin{tabular}[c]{@{}l@{}}\\$-$ \\ $-$ \\ $t(19) = -13.83$\end{tabular}
		&
		\begin{tabular}[c]{@{}l@{}}\\$-$ \\ $-$ \\ $p < .001^\text{***}$ \end{tabular}
		& 
		\begin{tabular}[c]{@{}l@{}}\\$M = 0.14, SD = 0.03$ \\ $M = 0.22, SD = 0.03$ \\ $-$\end{tabular}
		& 
		\begin{tabular}[c]{@{}l@{}}\\$-$ \\ $-$ \\ $t(19) = 11.96$\end{tabular}
		& 
		\begin{tabular}[c]{@{}l@{}}\\$-$ \\ $-$ \\ $p < .001^\text{***}$\end{tabular}
		\\  \hline
		\begin{tabular}[c]{@{}l@{}}\textit{Scenario 2} \\ MLOII \\ BOIR \\ MLOII vs BOIR \end{tabular}
		&
		\begin{tabular}[c]{@{}l@{}}\\$M = 88.90, SD = 9.47$ \\ $M = 94.00, SD = 7.85$ \\ $-$ \end{tabular}
		& 
		\begin{tabular}[c]{@{}l@{}}\\$-$ \\ $-$ \\ $Z = -2.20$\end{tabular}
		&
		\begin{tabular}[c]{@{}l@{}}\\$-$ \\ $-$ \\ $p = .03^\text{*}$ \end{tabular}
		& 
		\begin{tabular}[c]{@{}l@{}}\\$M = 0.10, SD = 0.05$ \\ $M = 0.13, SD = 0.02$ \\ $-$\end{tabular}
		& 
		\begin{tabular}[c]{@{}l@{}}\\$-$ \\ $-$ \\ $Z = -3.024$\end{tabular}
		& 
		\begin{tabular}[c]{@{}l@{}}\\$-$ \\ $-$ \\ $p = .002^\text{**}$\end{tabular}
		\\\hline
		\begin{tabular}[c]{@{}l@{}}\textit{Scenario 3} \\ MLOII \\ BOIR \\ MLOII vs BOIR \end{tabular}
		&
		\begin{tabular}[c]{@{}l@{}}\\$M = 67.70, SD = 2.49$ \\ $M = 67.65, SD = 2.03$ \\ $-$ \end{tabular}
		& 
		\begin{tabular}[c]{@{}l@{}}\\$-$ \\ $-$ \\ $t(19) = -0.093$\end{tabular}
		&
		\begin{tabular}[c]{@{}l@{}}\\$-$ \\ $-$ \\ $p = .927^\text{N.S.}$ \end{tabular}
		& 
		\begin{tabular}[c]{@{}l@{}}\\$M = 0.32, SD = 0.02$ \\ $M = 0.35, SD = 0.02$ \\ $-$\end{tabular}
		& 
		\begin{tabular}[c]{@{}l@{}}\\$-$ \\ $-$ \\ $t(19) = 4.095$\end{tabular}
		& 
		\begin{tabular}[c]{@{}l@{}}\\$-$ \\ $-$ \\ $p < .001^\text{***}$\end{tabular}
		\\ \hline
		\begin{tabular}[c]{@{}l@{}}\textit{Scenario 4} \\ MLOII \\ BOIR \\ MLOII vs BOIR \end{tabular}
		&
		\begin{tabular}[c]{@{}l@{}}\\$M = 44.85, SD = 17.73$ \\ $M = 65.00, SD = 14.72$ \\ $-$ \end{tabular}
		& 
		\begin{tabular}[c]{@{}l@{}}\\$-$ \\ $-$ \\ $t(19) = 2.981$\end{tabular}
		&
		\begin{tabular}[c]{@{}l@{}}\\$-$ \\ $-$ \\ $p = .008^\text{**}$ \end{tabular}
		& 
		\begin{tabular}[c]{@{}l@{}}\\$M = 0.51, SD = 0.10$ \\ $M = 0.44, SD = 0.15$ \\ $-$\end{tabular}
		& 
		\begin{tabular}[c]{@{}l@{}}\\$-$ \\ $-$ \\ $Z = -1.792$\end{tabular}
		& 
		\begin{tabular}[c]{@{}l@{}}\\$-$ \\ $-$ \\ $p = .073^\text{N.S.}$\end{tabular}
		\\ \hline
    \multicolumn{7}{@{}l}{\footnotesize \begin{tabular}{@{}l@{}}
    \rule{0pt}{9pt}$^\text{N.S.} \text{Non-Significant}$, $^\text{*} p < 0.05$, $^\text{**} p < 0.01$, $^\text{***} p < 0.001$
    \end{tabular}}
	    \end{tabular}
	\label{table:statistical_analysis}
\end{table*}

\subsection{Experimental Protocol}
We evaluated the MLOII using recorded\footnote{robot sensor data collected from a previous experiment's ROSbag log files.} data from a previous experiment with human operators, conducted in \cite{BOIR}. The latter allowed us to recreate the original experimental set-up and test our new hypotheses based on the same replayed data, enabling direct comparison and reproducibility.

The data was collected from 4 participants with a mean age of $M = 28.5$ $(SD = 1.41)$, all of whom had operated similar robotic systems extensively in the past and were either experts at utilizing the OCU and the GUI or had a lot of experience doing so. Each participant performed five trials on each of the four scenarios, resulting in a total of 20 trials for each one of them. The four scenarios used for experimental validation are briefly described below.

\textbf{Scenario 1} is the simplest one, in which the operators had to move from the starting position to target \say{b} and switch to \say{a} midway along (see Fig. \ref{fig:area1}).

\textbf{Scenario 2}'s map layout is more complex and the participants were forced to navigate around obstacles in order to access the intended goal. The participant's intent in each trial was to get to goal \say{c} (see Fig. \ref{fig:area2}).

\textbf{Scenario 3} evaluates the system's performance in a sequential goal configuration, where the participants were instructed to sequentially navigate through three sub-areas from the starting position to \say{a}, then \say{b}, and finally \say{c} goal (see Fig. \ref{fig:area3}).

\textbf{Scenario 4} accounts for assessing the performance of the system on how it adapts to an increased number of potential goals (here a total of five goals) spread across a large area (see Fig. \ref{fig:area4}). The participants were instructed to navigate between two POIs, randomly selected at the beginning of each trial.

\subsection{Performance metrics \& Comparisons}
The proposed MLOII's performance is compared to the algorithm proposed in \cite{BOIR} named Bayesian Operator Intent Recognition (BOIR). Specifically, BOIR utilizes probabilistic estimation for navigational human-robot teaming tasks and has been demonstrated to outperform other approaches. To assess the effectiveness of the proposed method and comparison purposes, the following metrics are taken into account: 1) \textit{accuracy} (i.e., the percentage of correctly predicted outcomes); b) \textit{uncertainty} (i.e., the divergence between the predicted and the actual value using the cross-entropy (log-loss) function \cite{Goodfellow-et-al-2016}) across each trial, and formally defined as follows:

\begin{equation}\label{accuracy}
\text{Accuracy} = \frac{\text{Number of correct predictions}}{\text{Total number of predictions}}
\end{equation}

\begin{equation}\label{uncertainty}
\text{Uncertainty}=L(y, \hat{y}) = - \frac{1}{N}\sum_{i=1}^{N} y_{i}\log(\hat{y}_{i})
\end{equation}
where, $y$ is the actual label, $\hat{y}$ is the predicted probability given by the classifier's output and $N$ is the total number of hypotheses made (i.e., total number of goals).

\begin{figure}
	\centering
	\includegraphics[width=0.95\columnwidth]{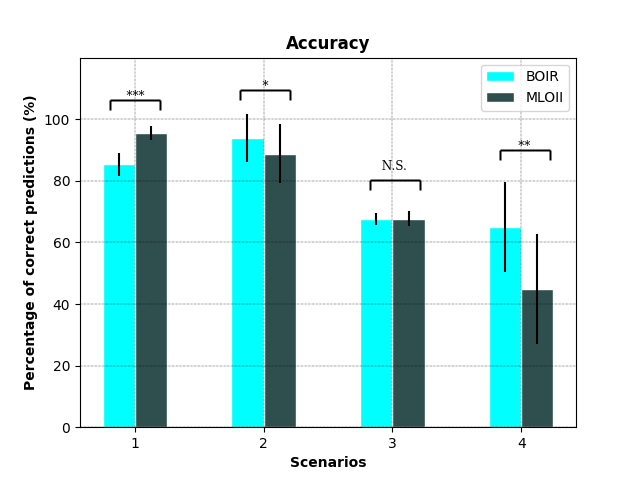}
	\caption{Accuracy (i.e., the percentage of correct predictions from the total predictions made) performance comparison across each method in every task scenario. Plot shows the means and standard deviations along with levels of significance.}
	\label{fig:accuracy_plot}
\end{figure}

\begin{figure}
	\centering
	\includegraphics[width=0.95\columnwidth]{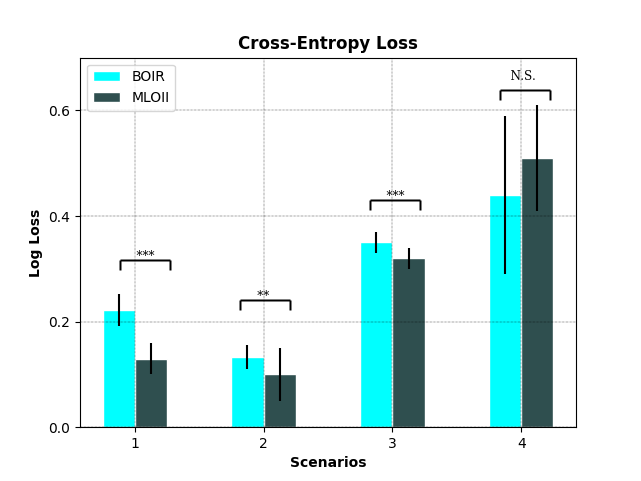}
	\caption{Log-loss performance comparison across each method in every task scenario along with levels of significance. The log-loss metric measures the uncertainty of predictions by penalizing those predictions that are confident and wrong. A perfect model would have a log-loss of 0.} 
	\label{fig:logloss_plot}
\end{figure}

\section{Results}
Table \ref{table:statistical_analysis} summarizes the experimental findings, including descriptive statistics and pairwise comparisons, with the former to be also illustrated graphically in Fig. \ref{fig:accuracy_plot} and \ref{fig:logloss_plot}. Paired sample t-tests were used for data conforming to a normal distribution according to the Shapiro–Wilk test, while the non-parametric Wilcoxon signed-rank test was used for the remaining data. Results were considered statistically significant if $p<.05$.

The results show that the MLOII outperforms (statistically significant) BOIR with regard to uncertainty across the three out of four scenarios. In particular, the MLOII method in Scenarios 1-3 predicts the most probable goal with more confidence (i.e., low uncertainty) as opposed to BOIR with an average of 0.19 and 0.23, respectively. In Scenario 4, the MLOII performs less effectively (0.51) compared to BOIR (0.44) without a statistically significant difference to be observed. In terms of accuracy, the performance of both methods seems to be equally distributed across three out of four scenarios. Specifically, in Scenario 1, the MLOII demonstrates accuracy higher by $10.5\%$ (statistically significant) compared to BOIR, whereas, in Scenario 2 BOIR outperforms (statistically significant) the MLOII by $5\%$. No statistically significant difference is reported between the two methods' accuracy in Scenario 3, with both approaches quantitatively performing at the same level of around $68\%$. Lastly, in Scenario 4 BOIR outperforms (statistically significant) the MLOII by approximately $15\%$.

\section{Discussion}
Results demonstrate that the proposed MLOII performs effectively at predicting the human intent across various navigational scenarios. First, it appears that the MLOII contributes to a statistically significant improvement in accuracy performance while operating in the simplest case scenarios (Scenario 1). This may be attributed to the use of more features as opposed to BOIR, to infer the navigational intent more accurately. Specifically, three features (i.e. approach velocity, Euclidean distance, angle) are adopted by the MLOII against two (i.e. path length, angle) by BOIR.

%uncertainty's performance along scenarios
Furthermore, the MLOII's capability of consistently sustaining lower uncertainty when compared to BOIR across most of the scenarios is worth mentioning. In other words, as each trial progresses, the predicted probability diverges less from the actual label. This could be assigned to the high confidence that the algorithm has once it gets the correct goal. 

% performance in more complex where MLOII underperforms
However, when it comes to dealing with scenarios in which the goals are arranged behind obstacles (Scenario 2) or distributed over a large area (Scenario 4), the MLOII's accuracy underperforms (statistically significant) against BOIR. This could be ascribed to the fact that no variety of map layouts were included in the training phase and that the data collected for generalization was limited. Another plausible explanation is the difference in the mathematical structure of the two algorithms. Contrary to BOIR, which is more easily integrable since no training phase is required, the MLOII cannot incrementally update the model by incorporating prior knowledge and the transition state belief. However, we expect that by modifying the MLOII to integrate Bayesian setting and utilize the path length in a fashion similar to BOIR could help improve performance in later scenarios.

% performance in sequentially located goals
Another interesting finding is that both methods perform identically regarding accuracy, especially when goals are adjusted in a straight line and sequentially accessed (Scenario 3). For instance, at the beginning of Scenario 3, both methods predict that the operator intends to move towards goal \say{b}, even though the initial goal needed to be inspected is \say{a}. Neither the number of features nor the mathematical structure adopted by these two methods seems to account for a better accuracy of results under these circumstances. This could be attributed to the difficulty of the AI agent to accurately predict the human intent in these cases without being given explicit knowledge (e.g., verbal communication or gesture cues), thus making the intent ambiguous and several interpretations plausible.

For future work, considering the MLOII's benefits, we propose three immediate directions. First, the MLOII has the advantage of potentially improving its performance, especially in cases where there are many goals. This could be achieved in addition to providing it with high quality data by increasing the size of its training data (e.g., training trials in multiple environments, operators with different driving behaviours). This would let the algorithm be scalable and generalize better to unseen environments. Second, the MLOII can be implemented into variable autonomy systems such as MI systems to tackle conflicts on transfer of control authority between the human operator and the AI agent in a similar way to \cite{hier}. Lastly, associated with the task of inferring the human navigational intent, an extension of the proposed method will be utilized for online identification and tracking of the potential POIs (e.g., hazards, fire, victims, radiation). Consequently, the AI agent will be able to help the human operators more actively by dynamically reconfiguring the surroundings and advising them on the best course of action.

% potential extension of the current discussion if we assumed direct comparison MLOII vs BOIR here
%BOIR advantages: 1) Incorporating prior knowledge or beliefs, 2) By using Bayesian updating, we can incrementally update the model, 3) no training needed contrast to ML techniques, 4) a rigorous method for inference, which can incorporate both data (in the likelihood) and theory (in the prior).

% MLOII advantages: 1) perfrom well on tabular data, 2) allow for bayesian integration into their structure, 3) less computation and mathematically (statistically) challenging contrast to Bayes.  

\section{Conclusion}
In this paper, we propose the MLOII algorithm to predict human operator intent in navigational tasks. Specifically, we use an offline learning process to train a Random Forest classifier to recognize the goal the operator is moving to by utilizing the robot's spatial properties (i.e., approach velocity, distance, angle). During run-time, we use the trained method to correctly infer the human's intended navigation goal. Moreover, an experimental study was carried out to evaluate the effectiveness of the system inspired by a remote disaster inspection task. The MLOII was tested and compared against another method from literature, on its performance across multiple scenarios using recorded data from previous human subject experiments. Findings suggest that the machine learning based methods are capable of accurately inferring the human operator's navigational intent with low uncertainty and can be used in the HRI context. Future work includes improving the training process, extending the current capabilities, and exploiting the presented inference method as a supplementary tool in order to improve HRI schemes such as MI systems.    

% \begin{ack}
% Place acknowledgments here.
% \end{ack}

\bibliography{ifacconf}             % bib file to produce the bibliography

\begin{thebibliography}{23}
\providecommand{\natexlab}[1]{#1}
\providecommand{\url}[1]{\texttt{#1}}
\providecommand{\urlprefix}{URL }
\expandafter\ifx\csname urlstyle\endcsname\relax
  \providecommand{\doi}[1]{doi:\discretionary{}{}{}#1}\else
  \providecommand{\doi}{doi:\discretionary{}{}{}\begingroup
  \urlstyle{rm}\Url}\fi

\bibitem[{Breiman(2001)}]{Breiman2001}
Breiman, L. (2001).
\newblock Random forests.
\newblock \emph{Machine Learning}, 45, 5--32.
\newblock \doi{10.1023/A:1010933404324}.

\bibitem[{Carlson and Demiris(2008)}]{Carlson2008}
Carlson, T. and Demiris, Y. (2008).
\newblock {Human-wheelchair collaboration through prediction of intention and
  adaptive assistance}.
\newblock \emph{Proceedings - IEEE International Conference on Robotics and
  Automation}, 3926--3931.
\newblock \doi{10.1109/ROBOT.2008.4543814}.

\bibitem[{Chatziparaschis et~al.(2020)Chatziparaschis, Lagoudakis, and
  Partsinevelos}]{Chatziparaschis}
Chatziparaschis, D., Lagoudakis, M.G., and Partsinevelos, P. (2020).
\newblock Aerial and ground robot collaboration for autonomous mapping in
  search and rescue missions.
\newblock \emph{Drones}, 4(4).
\newblock \doi{10.3390/drones4040079}.
\newblock \urlprefix\url{https://www.mdpi.com/2504-446X/4/4/79}.

\bibitem[{Chiou et~al.(2021)Chiou, Hawes, and Stolkin}]{chiouMI}
Chiou, M., Hawes, N., and Stolkin, R. (2021).
\newblock Mixed-initiative variable autonomy for remotely operated mobile
  robots.
\newblock \emph{J. Hum.-Robot Interact.}, 10(4).
\newblock \doi{10.1145/3472206}.

\bibitem[{Choi and Lee(2021)}]{CHOI}
Choi, D. and Lee, S. (2021).
\newblock Comparison of machine learning algorithms for predicting lane
  changing intent.
\newblock \emph{International journal of automotive technology}, 22(2),
  507--518.
\newblock \doi{10.1007/s12239-021-0047-x}.
\newblock \urlprefix\url{http://dx.doi.org/10.1007/s12239-021-0047-x}.

\bibitem[{Demiris(2007)}]{Demiris2007}
Demiris, Y. (2007).
\newblock Prediction of intent in robotics and multi-agent systems.
\newblock \emph{Cognitive processing}, 8(3), 151—158.
\newblock \doi{10.1007/s10339-007-0168-9}.
\newblock \urlprefix\url{https://doi.org/10.1007/s10339-007-0168-9}.

\bibitem[{Dragan and Srinivasa(2013)}]{Dragan2013}
Dragan, A.D. and Srinivasa, S.S. (2013).
\newblock {A policy-blending formalism for shared control}.
\newblock \emph{International Journal of Robotics Research}, 32(7), 790--805.
\newblock \doi{10.1177/0278364913490324}.

\bibitem[{Goodfellow et~al.(2016)Goodfellow, Bengio, and
  Courville}]{Goodfellow-et-al-2016}
Goodfellow, I., Bengio, Y., and Courville, A. (2016).
\newblock \emph{Deep Learning}.
\newblock MIT Press.
\newblock \url{http://www.deeplearningbook.org}.

\bibitem[{Grinsztajn et~al.(2022)Grinsztajn, Oyallon, and
  Varoquaux}]{https://doi.org/10.48550/arxiv.2207.08815}
Grinsztajn, L., Oyallon, E., and Varoquaux, G. (2022).
\newblock Why do tree-based models still outperform deep learning on tabular
  data?
\newblock \doi{10.48550/ARXIV.2207.08815}.
\newblock \urlprefix\url{https://arxiv.org/abs/2207.08815}.

\bibitem[{Huntemann et~al.(2013)Huntemann, Demeester, Poorten, {Van Brussel},
  and {De Schutter}}]{Huntemann2013}
Huntemann, A., Demeester, E., Poorten, E.V., {Van Brussel}, H., and {De
  Schutter}, J. (2013).
\newblock {Probabilistic approach to recognize local navigation plans by fusing
  past driving information with a personalized user model}.
\newblock \emph{Proceedings - IEEE International Conference on Robotics and
  Automation}, 4376--4383.
\newblock \doi{10.1109/ICRA.2013.6631197}.

\bibitem[{Jain and Argall(2019)}]{Jain2019}
Jain, S. and Argall, B. (2019).
\newblock {Probabilistic Human Intent Recognition for Shared Autonomy in
  Assistive Robotics}.
\newblock \emph{ACM Trans. Hum.-Robot Interact.}, 9.
\newblock \doi{10.1145/3359614}.

\bibitem[{Jiang and Arkin(2015)}]{Jiang2015}
Jiang, S. and Arkin, R.C. (2015).
\newblock {Mixed-Initiative Human-Robot Interaction: Definition, Taxonomy, and
  Survey}.
\newblock In \emph{IEEE International Conference on Systems, Man, and
  Cybernetics (SMC)}, 954--961.
\newblock \doi{10.1109/SMC.2015.174}.

\bibitem[{Liu et~al.(2016)Liu, Hamrick, Fisac, Dragan, Hedrick, Sastry, and
  Griffiths}]{Liu2016}
Liu, C., Hamrick, J.B., Fisac, J.F., Dragan, A.D., Hedrick, J.K., Sastry, S.S.,
  and Griffiths, T.L. (2016).
\newblock {Goal inference improves objective and perceived performance in
  human-robot collaboration}.
\newblock \emph{Proceedings of the International Joint Conference on Autonomous
  Agents and Multiagent Systems, AAMAS}, 940--948.

\bibitem[{Murphy and Schreckenghost(2013)}]{Murphy}
Murphy, R.R. and Schreckenghost, D. (2013).
\newblock Survey of metrics for human-robot interaction.
\newblock In \emph{2013 8th ACM/IEEE International Conference on Human-Robot
  Interaction (HRI)}, 197--198.
\newblock \doi{10.1109/HRI.2013.6483569}.

\bibitem[{Panagopoulos et~al.(2022)Panagopoulos, Petousakis, Ramesh, Ruan,
  Nikolaou, Stolkin, and Chiou}]{hier}
Panagopoulos, D., Petousakis, G., Ramesh, A., Ruan, T., Nikolaou, G., Stolkin,
  R., and Chiou, M. (2022).
\newblock A hierarchical variable autonomy mixed-initiative framework for
  human-robot teaming in mobile robotics.
\newblock In \emph{2022 IEEE 3rd International Conference on Human-Machine
  Systems (ICHMS)}, 1--6.
\newblock \doi{10.1109/ICHMS56717.2022.9980686}.

\bibitem[{Panagopoulos et~al.(2021)Panagopoulos, Petousakis, Stolkin, Nikolaou,
  and Chiou}]{BOIR}
Panagopoulos, D., Petousakis, G., Stolkin, R., Nikolaou, G., and Chiou, M.
  (2021).
\newblock A bayesian-based approach to human operator intent recognition in
  remote mobile robot navigation.
\newblock In \emph{2021 IEEE International Conference on Systems, Man, and
  Cybernetics (SMC)}, 125--131.
\newblock \doi{10.1109/SMC52423.2021.9658942}.

\bibitem[{Pappas et~al.(2020)Pappas, Chiou, Epsimos, Nikolaou, and
  Stolkin}]{pappas}
Pappas, P., Chiou, M., Epsimos, G.T., Nikolaou, G., and Stolkin, R. (2020).
\newblock Vfh+ based shared control for remotely operated mobile robots.
\newblock In \emph{2020 IEEE International Symposium on Safety, Security, and
  Rescue Robotics (SSRR)}, 366--373.
\newblock \doi{10.1109/SSRR50563.2020.9292585}.

\bibitem[{Petković et~al.(2019)Petković, Puljiz, Marković, and
  Hein}]{PETKOVIC2019182}
Petković, T., Puljiz, D., Marković, I., and Hein, B. (2019).
\newblock Human intention estimation based on hidden markov model motion
  validation for safe flexible robotized warehouses.
\newblock \emph{Robotics and Computer-Integrated Manufacturing}, 57, 182--196.
\newblock \doi{https://doi.org/10.1016/j.rcim.2018.11.004}.

\bibitem[{Petousakis et~al.(2020)Petousakis, Chiou, Nikolaou, and
  Stolkin}]{Petousakis2020}
Petousakis, G., Chiou, M., Nikolaou, G., and Stolkin, R. (2020).
\newblock {Human operator cognitive availability aware Mixed-Initiative
  control}.
\newblock In \emph{IEEE International Conference on Human-Machine Systems
  (ICHMS)}, 1--4. IEEE.
\newblock \doi{10.1109/ICHMS49158.2020.9209582}.

\bibitem[{Rognon et~al.(2018)Rognon, Mintchev, Dellagnola, Cherpillod, Atienza,
  and Floreano}]{Rognon2018}
Rognon, C., Mintchev, S., Dellagnola, F., Cherpillod, A., Atienza, D., and
  Floreano, D. (2018).
\newblock {FlyJacket: An Upper Body Soft Exoskeleton for Immersive Drone
  Control}.
\newblock \emph{IEEE Robotics and Automation Letters}, 3(3), 2362--2369.
\newblock \doi{10.1109/LRA.2018.2810955}.

\bibitem[{Rothfu{\ss} et~al.(2022)Rothfu{\ss}, Chiou, Inga, Hohmann, and
  Stolkin}]{Nemics}
Rothfu{\ss}, S., Chiou, M., Inga, J., Hohmann, S., and Stolkin, R. (2022).
\newblock A negotiation-theoretic framework for control authority transfer in
  mixed-initiative robotic systems.

\bibitem[{Yan et~al.(2019)Yan, Gao, Zhang, and Chang}]{YanLSTM}
Yan, L., Gao, X., Zhang, X., and Chang, S. (2019).
\newblock Human-robot collaboration by intention recognition using deep lstm
  neural network.
\newblock In \emph{2019 IEEE 8th International Conference on Fluid Power and
  Mechatronics (FPM)}, 1390--1396.
\newblock \doi{10.1109/FPM45753.2019.9035907}.

\bibitem[{Zolotas and Demiris(2021)}]{zolotas2021}
Zolotas, M. and Demiris, Y. (2021).
\newblock Disentangled sequence clustering for human intention inference.
\newblock \doi{10.48550/ARXIV.2101.09500}.
\newblock \urlprefix\url{https://arxiv.org/abs/2101.09500}.

\end{thebibliography}
                                                     % with bibtex (preferred)
                                                   
%\begin{thebibliography}{xx}  % you can also add the bibliography by hand

%\bibitem[Able(1956)]{Abl:56}
%B.C. Able.
%\newblock Nucleic acid content of microscope.
%\newblock \emph{Nature}, 135:\penalty0 7--9, 1956.

%\bibitem[Able et~al.(1954)Able, Tagg, and Rush]{AbTaRu:54}
%B.C. Able, R.A. Tagg, and M.~Rush.
%\newblock Enzyme-catalyzed cellular transanimations.
%\newblock In A.F. Round, editor, \emph{Advances in Enzymology}, volume~2, pages
%  125--247. Academic Press, New York, 3rd edition, 1954.

%\bibitem[Keohane(1958)]{Keo:58}
%\newblock \emph{Power and Interdependence: World Politics in Transitions}.
%\newblock Little, Brown \& Co., Boston, 1958.

%\bibitem[Powers(1985)]{Pow:85}
%\newblock Is there a way out?
%\newblock \emph{Harpers}, pages 35--47, June 1985.

%\bibitem[Soukhanov(1992)]{Heritage:92}
%\newblock \emph{{The American Heritage. Dictionary of the American Language}}.
%\newblock Houghton Mifflin Company, 1992.

%\end{thebibliography}

\end{document}